\documentclass[10pt,twocolumn,letterpaper]{article}

\usepackage{cvpr}              

\usepackage{graphicx}
\usepackage{amsmath}
\usepackage{amssymb}
\usepackage{booktabs}
\usepackage{xcolor}         
\usepackage{tabularx}
\usepackage{makecell}
\usepackage{float}
\usepackage[accsupp]{axessibility}

\usepackage[pagebackref,breaklinks,colorlinks]{hyperref}

\usepackage[capitalize]{cleveref}
\usepackage{multirow}
\crefname{section}{Sec.}{Secs.}
\Crefname{section}{Section}{Sections}
\Crefname{table}{Table}{Tables}
\crefname{table}{Tab.}{Tabs.}

\def\cvprPaperID{4197} 
\def\confName{CVPR}
\def\confYear{2022}

\begin{document}

\title{Group R-CNN for Weakly Semi-supervised Object Detection with Points }

\author{Shilong Zhang\textsuperscript{\rm 1}\thanks{Equal contribution} ,
         Zhuoran Yu\textsuperscript{\rm 2}\footnotemark[1] , 
        Liyang Liu\textsuperscript{\rm 3}\footnotemark[1] ,
       Xinjiang Wang\textsuperscript{\rm 4},
        Aojun Zhou\textsuperscript{\rm 4},
        Kai Chen \textsuperscript{\rm 1,4}\\
        \small
        \textsuperscript{\rm 1}Shanghai AI Laboratory  \space \space
    \small\textsuperscript{\rm 2} Georgia Institute of Technology \\
    \small\textsuperscript{\rm 3} Tencent AI Platform Department, China \space \space
        \small\textsuperscript{\rm 4} SenseTime Research \\
    \tt\small zhangshilong@pjlab.org.cn, zhuoranyu@gatech.edu, leonlyliu@tencent.com, \\
     \tt\small        \{wangxinjiang,zhouaojun,chenkai\}@sensetime.com}

\maketitle

\newcommand{\ProposedMethodName}{{{{Group R-CNN}}}}

\newcommand{\zhuoran}[1]{{\color{red}{(Zhuoran: #1)}}}

\begin{abstract}
We study the problem of weakly semi-supervised object detection with points (WSSOD-P), where the training data is combined by a small set of fully annotated images with bounding boxes and a large set of weakly-labeled images with only a single point annotated for each instance. The core of this task is to train a point-to-box regressor on well-labeled images that can be used to predict credible bounding boxes for each point annotation. We challenge the prior belief that existing CNN-based detectors are not compatible with this task. Based on the classic R-CNN architecture, we propose an effective point-to-box regressor: Group R-CNN. Group R-CNN first uses \textbf{instance-level proposal grouping} to generate a group of proposals for each point annotation and thus can obtain a high recall rate. To better distinguish different instances and improve precision, we propose instance-level proposal assignment to replace the vanilla assignment strategy adopted in original R-CNN methods. As naive instance-level assignment brings converging difficulty, we propose \textbf{instance-aware representation learning} which consists of instance-aware feature enhancement and instance-aware parameter generation to overcome this issue. Comprehensive experiments on the MS-COCO benchmark demonstrate the effectiveness of our method. Specifically, Group R-CNN significantly outperforms the prior method Point DETR by 3.9 mAP with 5\% well-labeled images, which is the most challenging scenario. The source code can be found at \url{https://github.com/jshilong/GroupRCNN}
\end{abstract}

\section{Introduction}
\label{sec:intro}

Object detection has witnessed great improvements with the development of network architectures and dataset construction in the past few years. However, advanced object detectors usually require training on a large number of images with accurate bounding box annotations, which are very time-consuming and expensive to obtain. To relieve the burden of human labeling, previous studies propose weakly-~\cite{li2019weakly, shi2017weakly, bilen2016weakly, jie2017deep, zhu2017soft} and
semi-supervised methods~\cite{gao2019note, tang2016large, zhou2021instant, liu2021unbiased} for object detection, which adopts a small portion of well-annotated images alongside abundant weakly-annotated images and unlabeled images, respectively. Weakly semi-supervised object detection (WSSOD)~\cite{yan2017weakly} takes goods from both worlds. By replacing the unlabeled images in semi-supervised detection with weakly-labeled images, it achieves a good balance between labeling costs and model performance. One step further, WSSOD with point annotations (WSSOD-P) labels an instance with a single point so as to provide both category and location information with minimal labeling cost. The time cost of labeling points is comparable with that of providing image-level annotations~\cite{bearman2016s, chen2021points}, which is substantially lower than bounding box annotations.

\begin{figure}[t]
    \centering
    \includegraphics[width=\linewidth]{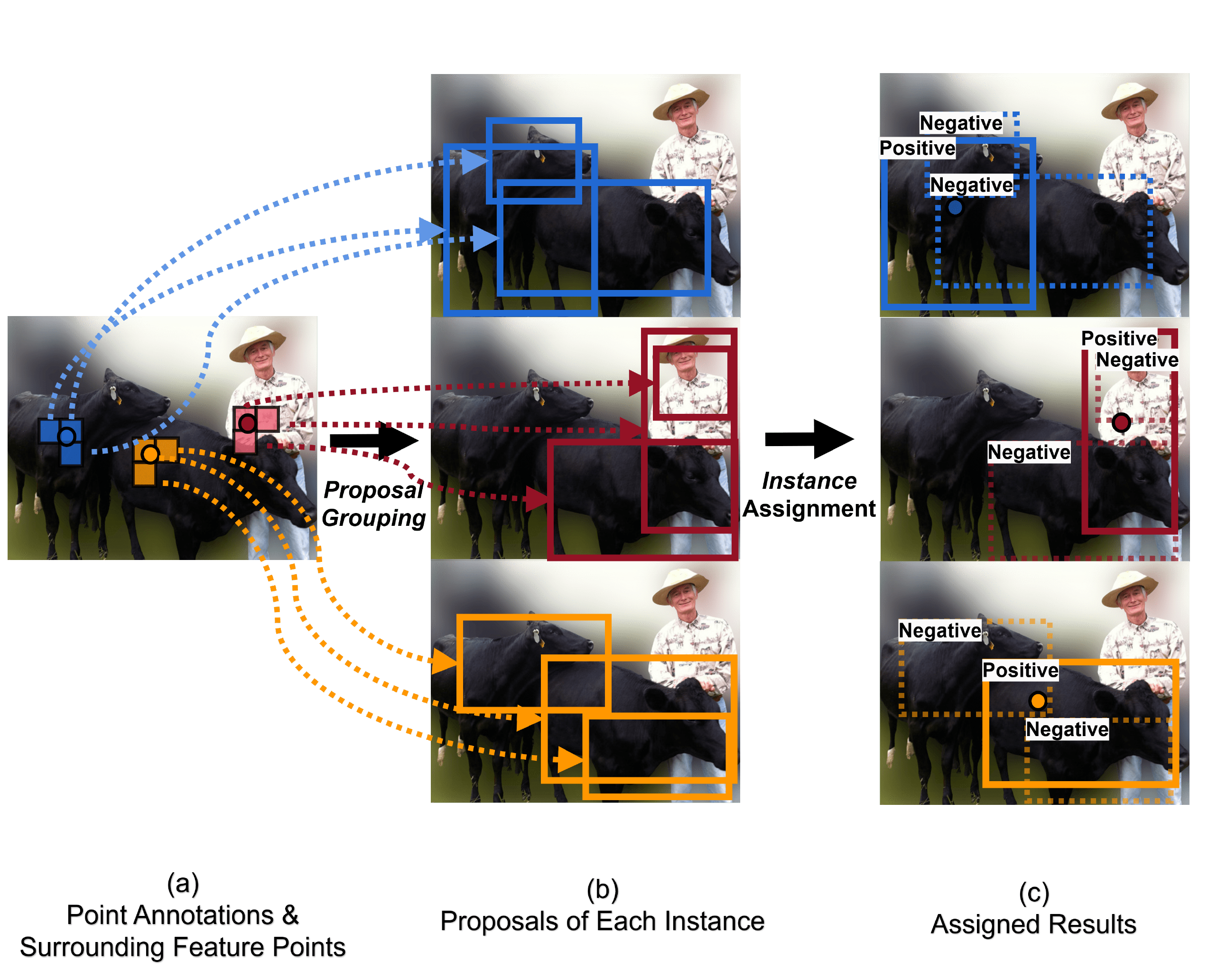}
    \caption{
    \textbf{An illustration of instance-level proposal grouping and instance-level proposal assignment}. Proposals in the same group (i.e. belonging to the same instance) are denoted by the same color. (a) to (b): \textbf{Instance-level grouping}: rather than just only the annotated point, proposals generated by all points that are close enough to the (projected) annotation point are collected to form a group for each instance. (b) to (c): \textbf{Instance-level assignment}: A proposal is assigned as positive if and only if the IoU with its \textit{belonging} instance is above a pre-defined threshold, regardless of its IoU with any other instances.
    }
    \label{fig:intro}
    \vspace{-3mm}

\end{figure}

Prior work~\cite{chen2021points} proposes the following pipeline to leverage point annotations. \textbf{(1)} train a point-to-box regressor using images with only bounding box annotations. To simulate the process of translating points into bounding boxes, points are randomly sampled within bounding boxes of instances as point annotations. \textbf{(2)} After training, the regressor is used to transform point annotations on weakly-labeled images to pseudo bounding box annotations. \textbf{(3)} Finally, any object detector can be trained with both well-labeled images and pseudo-labeled images in a supervised fashion. The core of this task then boils down to designing an accurate point-to-box regressor.

Point DETR~\cite{chen2021points} claims that the CNN-based detector performs poorly as a point-to-box regressor, but we argue it does not hold true based on the following in-depth analysis. On one hand, we find that CNN models in prior work only generate a single proposal with the feature vector \textbf{located at the annotated point} (projected on the feature map). The quality of such a proposal can be quite low, resulting in a low recall rate since annotated points can deviate from object centers and are less informative for box regression. To improve the recall rate with arbitrarily annotated points, we propose instance-level proposal grouping (Figure \ref{fig:intro}, (a) to (b)) that generates a group of proposals rather than a single proposal for each instance. That said, points on feature maps that are \textbf{close to} the annotated point will be taken into consideration, and all proposals generated from these points form an instance-level group for a certain instance.

On the other hand, the original proposal assignment strategy assigns a proposal to be positive if its maximal IoU with \textbf{any ground truth boxes} surpasses a certain threshold. However, proposals in one group may be assigned to the ground-truth bounding box of another group when objects with the same category are close to each other (e.g. in the crowded scene). Such a scenario is very common in real-world datasets, for example, in MS-COCO over 50\% images have such a scene. Training with such a strategy cannot ensure a high-quality proposal is produced for each point annotation. Therefore, we propose instance-level assignment that assigns proposals in a group only to its \textbf{belonging ground-truth box}. A proposal will be treated as negative if its IoU with the corresponding instance is low, regardless that its IoU with another instance may surpass the threshold (Figure \ref{fig:intro}, (b) to (c)).

Nevertheless, we find that the naive instance assignment causes converging difficulty and thus leads to inferior performance than the original assignment strategy. Since proposals from different  groups share \textbf{the same feature map} and are also convolved with \textbf{the same fixed parameters}, positive proposals in one group and negative proposals in another group can have similar convolution outputs. However, they are assigned to completely opposite optimization targets, this contradiction can hinder model convergence. To tackle this problem, we propose instance-aware representation learning, consisting of instance-aware feature enhancement and instance-aware parameter generation. Specifically, we use point annotations to compute instance-aware relative coordinates, with which to construct instance-aware feature maps. To further distinguish features of proposals from different groups, we use both instance-aware features and the annotated points' category embeddings to generate instance-aware model parameters. By convolving the \textbf{instance-aware features} and the \textbf{instance-aware parameters}, we successfully mitigate the converging issue brought by naive instance assignment and obtain superior performance.

To highlight our key designs and the proposed general framework as a point-to-box regressor, we name our regressor Group R-CNN. Compared with the previous state-of-the-art method Point DETR which is based on the transformer architecture, our proposed Group R-CNN has the following advantages: (1) Group R-CNN can take advantage of the feature pyramid network~\cite{lin2017feature} for multi-scale proposal generation, while Point DETR~\cite{chen2021points} can not naively adopt FPN. (2) Group R-CNN inherits the convergence superiority from CNNs, and thus converges better and faster than Point DETR, especially under the low well-labeled data regime~\cite{dosovitskiy2020image} (See Appendix for details).

We conduct extensive experiments on MS-COCO dataset, using various percentages of labeled images to showcase the effectiveness of Group R-CNN. Group R-CNN outperforms the existing transformer-based method Point DETR by a large margin under different experimental protocols while only requiring half of the training budget. To be specific, our Group R-CNN outperforms Point DETR by 3.9 mAP when only 5\% images are well-labeled by bounding boxes. Thus, Group R-CNN achieves both better performance and faster convergence over previous state-of-the-art.

\begin{figure*}[htp!]
    \centering
    \includegraphics[width=\linewidth]{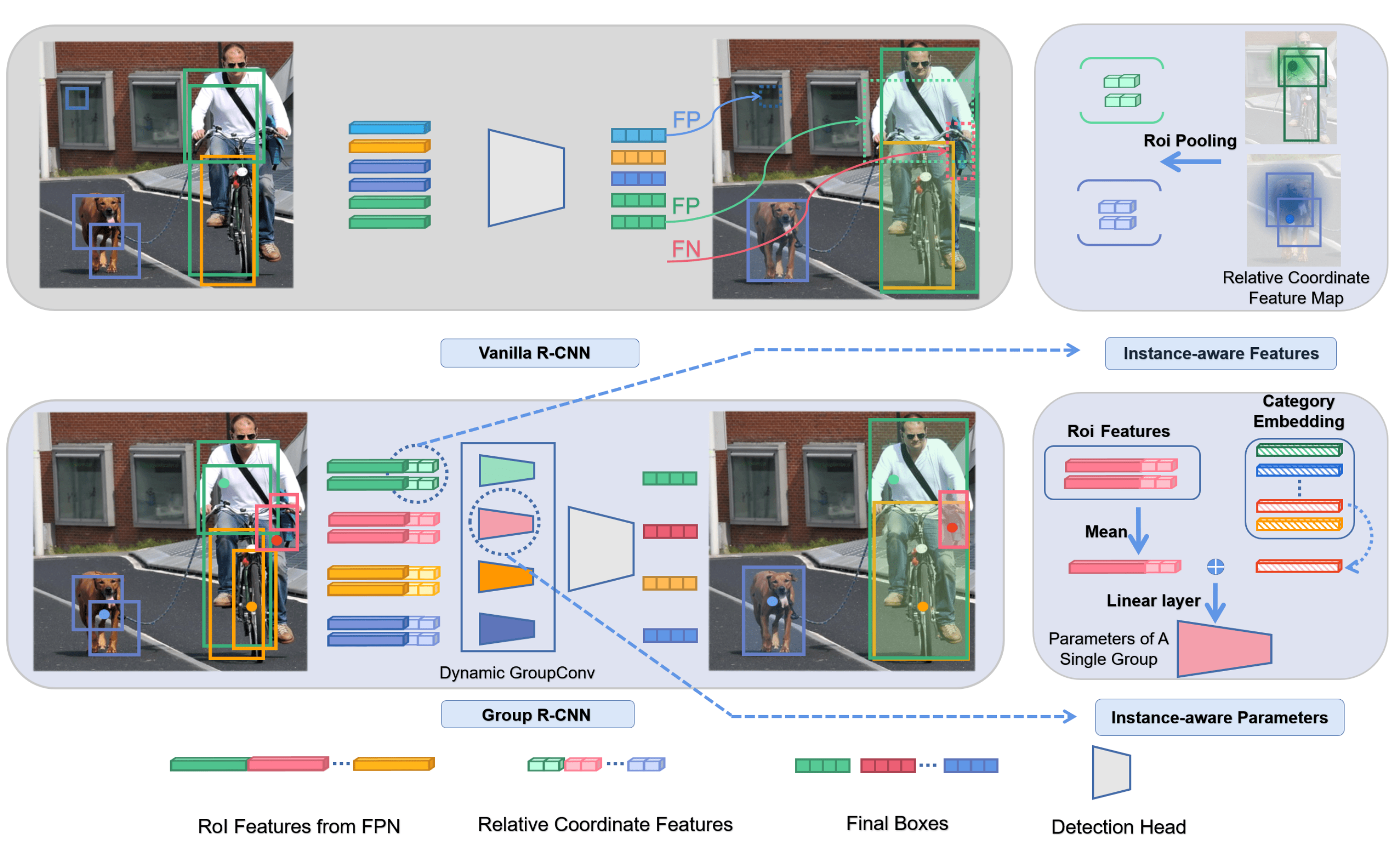}
    \caption{
    \textbf{The pipeline of Group R-CNN}. The Vanilla R-CNN (\textbf{the upper architecture}) cannot leverage the point annotation and easily produces false positive and false negative bounding boxes. In contrast, Group R-CNN (\textbf{the lower architecture}) achieves higher recall and precision by incorporating relative coordinates (top-right) and dynamic group convolution (bottom-right) for the point-to-box translation task. \textbf{Instance-aware Features (top-right)}: a separate feature map is constructed for each instance to encode the relative coordinates w.r.t. its point annotation, followed by the RoI-pooling procedure. \textbf{Instance-aware Parameters (bottom-right)}: the average RoI feature of each group is concatenated with the corresponding category embedding, acting as the input of a linear layer to generate parameters for the dynamic group convolution.
    }
    \label{fig:pipeline}

\end{figure*}

\section{Related Work}
\textbf{Supervised Object Detection.} Supervised object detection has made remarkable progress over the past few years~\cite{ren2015faster, cai2018cascade, law2018cornernet, lin2017focal, tian2019fcos}. Based on the architecture design, these object detectors can be mainly categorized as two-stage detectors and single-stage detectors. Two-stage detectors~\cite{ren2015faster} first generate a large set of object proposals and refine them in the second stage. In contrast, single-stage detectors~\cite{redmon2017yolo9000, lin2017focal, liu2016ssd} directly predicts bounding boxes with categories without refinement. In addition to designs of object detectors, multi-scale object detection is another popular line of work. Feature Pyramid Network~\cite{lin2017feature} produces multi-scale feature representation that can be used by both single-stage and two-stage object detectors

\textbf{Semi- and Weakly-Supervised Object Detection.} To reduce the annotation cost of producing large-scale well-labeled images for object detection, two popular settings were proposed. Semi-supervised object detection~\cite{liu2021unbiased, sohn2020simple, zhou2021instant} replaces the weakly-labeled images with unlabeled images and leverages pseudo-labeling for better performance. The problem considered in this work is different from these two classic learning paradigm as the point annotation is added to weakly-labeled images.  Weakly-supervised object detection~\cite{li2019weakly, shi2017weakly, yan2017weakly, bilen2016weakly, tang2018pcl, gao2021ts, wan2019c} proposes to use a large set of weakly-labeled images (such as image-level annotations without bounding boxes) alongside a small set of well-labeled images. PCL~\cite{tang2018pcl} incorporates a concept of grouping and clusters proposals in an unsupervised manner to prevent from focusing on discriminative parts of instances. In contrast, Group R-CNN selects proposals around the human-labeled points to \textbf{improve the recall rate}, which is different from PCL in terms of motivation and implementation. Details of our method can be found in Section ~\ref{sec:method}.

\textbf{Weakly Semi-Supervised Object Detection with Points (WSSOD-P).} Recently, Point DETR~\cite{chen2021points} proposes a new annotation format for weakly-supervised object detection, which adds point annotations as a new form of weak annotations. To leverage point annotations, Point DETR introduces a transformer-based point-to-box regressor to transform point annotations to bounding-box annotations so that an actual object detector can be trained. However, transformer-based models usually have convergence issues when training data is not sufficient~\cite{dosovitskiy2020image}, which is usually the case of WSSOD-P (See Appendix for details). In contrast, we design a convolution-based regressor for this task, which is considered challenging by prior work~\cite{chen2021points}.

\textbf{Dynamic Parameter Generation.}
Dynamic Parameter Generation has been used in various computer vision tasks such as image classification~\cite{yang2019condconv, chen2020dynamic, zhang2020dynet} and instance segmentation~\cite{tian2020conditional}. The key idea behind this line of work is to dynamically adjust network parameters based on inputs. Our proposed dynamic group convolution is inspired by these existing works and we propose to generate convolution parameters for each instance-level proposal groups. To the best of our knowledge, this is the first time such an idea is implemented under WSSOD-P setting.
\label{sec:related}

\section{Group R-CNN}
\label{sec:method}
In this section, we first review the problem of weakly semi-supervised object detection with point annotations (WSSOD-P). Next, we present our novel framework Group R-CNN as a solution to this task.

\subsection{Background}
The problem of WSSOD-P proposes to train an object detector using a small portion of \textbf{well-labeled} images with instance-level annotations (e.g., bounding boxes and class labels) and an abundant \textbf{weakly-labeled} images with only one single point annotation for each instance. Compared with vanilla weakly-supervised object detection, the point annotation provides meaningful location information for instances without introducing much labeling costs. A common pipeline \cite{chen2021points} to solve this problem is: (1) training a point-to-box regressor using well-labeled images with bounding box annotations (2) generating pseudo bounding boxes for images with point annotations, and (3) training an object detector with the combination of well-labeled and pseudo-labeled images. The core of this task is to design an effective point-to-box regressor that translates point annotations to credible pseudo bounding box annotations. 

To design a better point-to-box regressor, we propose Group R-CNN, a CNN-based architecture for this point-to-box translation task. Group R-CNN inherits a multi-stage architecture as in Cascade R-CNN~\cite{cai2018cascade}, which consists of a proposal generation stage and a proposal refinement stage. 
We introduce our novel designs of our architecture in detail in the next sections. 

\subsection{Instance-level Proposal Grouping}

\begin{figure}
    \centering
    \includegraphics[width=\linewidth]{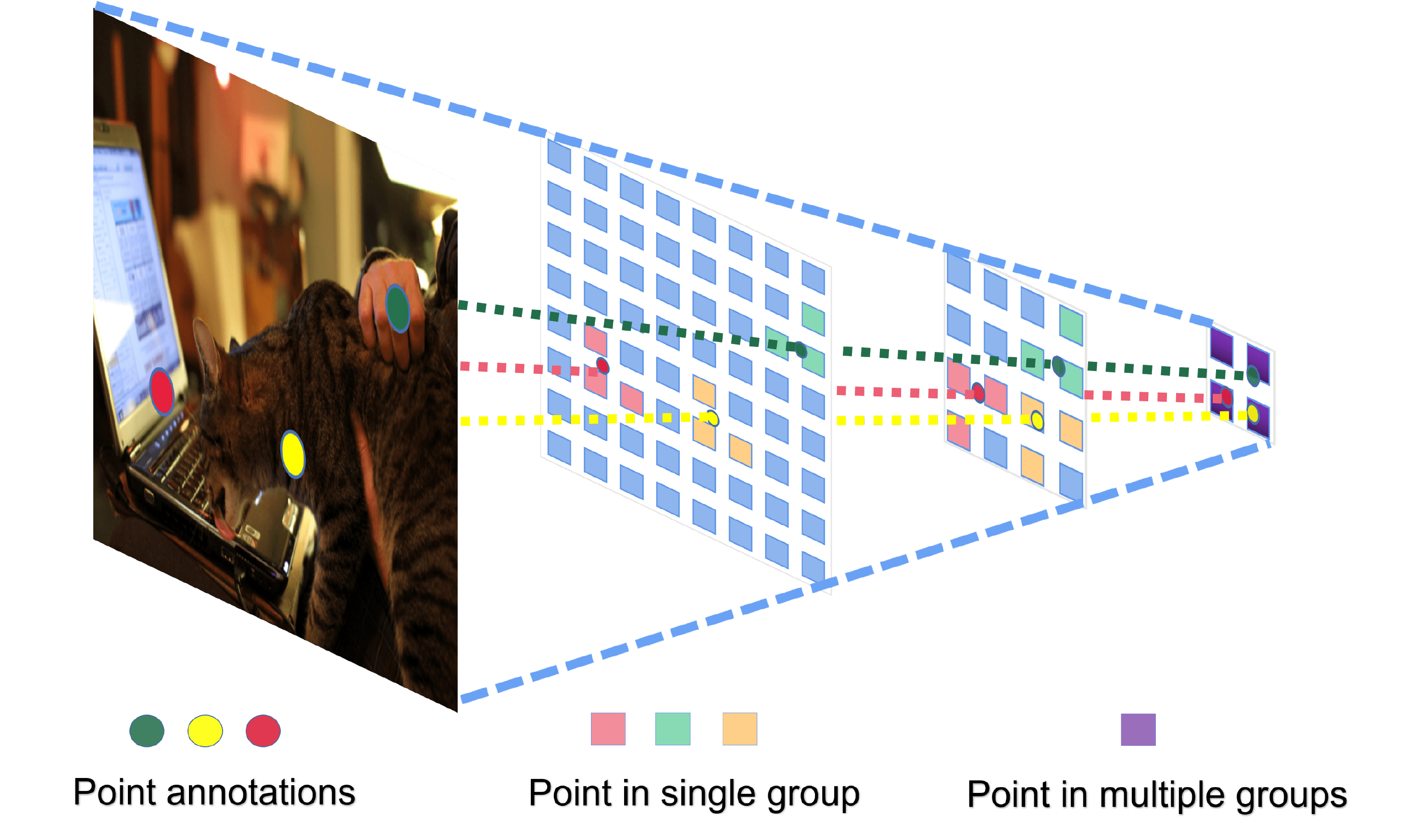}
    \caption{
    \textbf{Instance-level Proposal Grouping}. Proposals generated by points that are adjacent to projected annotation points form a single group. Such points can present in multiple groups.
    }
    \label{fig:instance_group}

\end{figure}

Previous attempts of using a CNN-based framework as point-to-box regressors  generate a single proposal using the feature of the projected annotation point for an instance, which may lead to a inferior recall rate resulting from inaccurate point annotations. To boost the recall rate, we propose instance-level proposal grouping (Figure ~\ref{fig:instance_group}), which aggregates proposals generated by feature points close to a certain annotation point to form a group. This strategy is based on the insight that points on feature maps corresponding to high-quality proposals are often close to the projection of annotation points on feature maps. Specifically, we collect $k$ feature points around the projected annotation point on each level of the feature pyramid. Proposals produced by these $km$ points then form a group for a given instance, where $m$ is the number of levels in FPN. Each group finally produces $nkm$ proposals where $n$ is the number of anchors with different scales and aspect ratios at each point in RPN. The proposed instance-level proposal grouping collects information from neighboring points around the annotated point and thus can improve recall rate and is more robust to the inaccurate annotation points.

\subsection{Instance-aware Representation Learning}

\textbf{Instance-level Proposal Assignment}. To achieve high recall while retraining high precision, the regressor is supposed to produce a single accurate bounding box for each point annotation. Therefore, we only output one proposal with the highest prediction score in a single group. However, the vanilla designs in classic R-CNN architectures have difficulty achieving such a goal especially when \textbf{objects with the same category are adjacent to each other} (crowded scene problem). First, in training, with the vanilla assignment strategy,  proposals are regarded as positive if their maximal IoU with \textbf{any ground truth boxes} is greater than a pre-defined threshold and are considered as negatives otherwise. However, when objects of the same class are close to each other, proposals within one group may be matched to the ground-truth bounding box of another instance and eventually harm the precision. To tackle this problem, we propose instance-level assignment so that proposals of a certain group can only be assigned to the corresponding ground-truth bounding box of this group. Proposals are treated as negative if the IoU between them and the corresponding instance is below the threshold even though they might reach the IoU threshold with ground-truth bounding boxes of another group. 

\begin{figure}[t]
    \centering
    \includegraphics[scale=0.28]{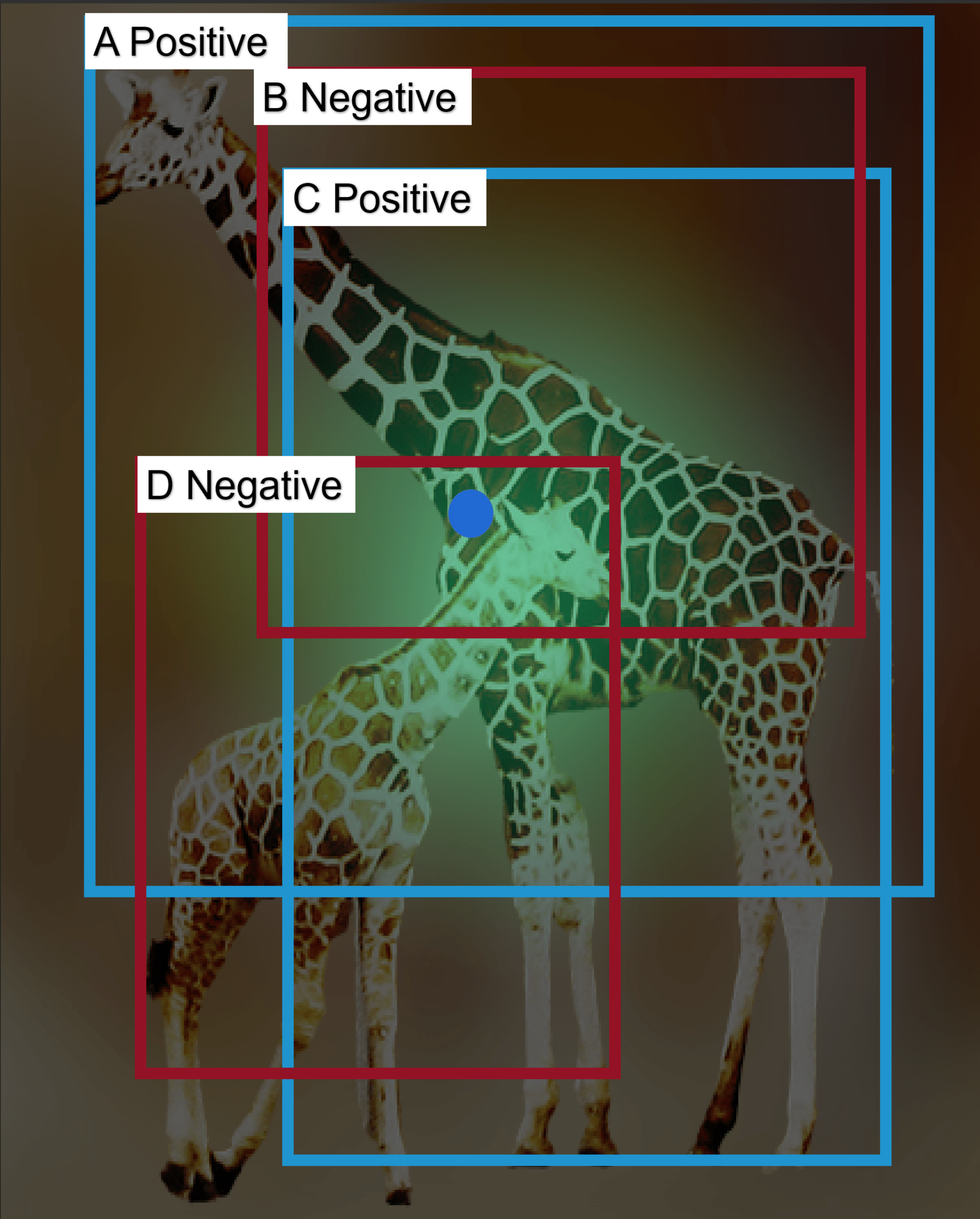}
    \caption{
    \textbf{An illustration of overlapped instances with the same category}. All four proposals are in the group of the taller giraffe. With the instance-level assignment, only proposal A and C are treated as positive. However, proposal D exposes a high-level similarity with positives of the group of the baby giraffe. \textbf{Because proposals in different groups share the feature pyramid and shared network parameters are used to process these features}, the predictions of positive proposals in the group of the baby giraffe and negative proposals in the group of tall giraffes are similar to each other, which causes confusion in model training.
    }
    \label{fig:limitaion}
\end{figure}

However, the crowded scene problem cannot be solved by naive instance-level assignment (as in Table \ref{tab:ablation_inst}). Since proposals in different groups share the same feature pyramid and the network also uses the shared parameters of R-CNN heads to process these features, positive proposals in one group and negative proposals in another group can have similar processed results, but are assigned to completely opposite optimization targets (as in Figure \ref{fig:limitaion}), which causes difficulty in model training. To address this difficulty, we propose our solutions from two different aspects: \textbf{instance-aware feature enhancement} through relative coordinates encoding and \textbf{instance-aware parameters} for a dynamic group convolution.

    

\begin{table}[htp!]
    \centering
                \caption{Failure of the naive instance-level assignment. Both methods are equipped \textbf{with instance-grouping} and output the best proposal of highest prediction score after NMS for each group.}
            \begin{tabular}{c|c|c|c}
            \hline
             & mAP & AP@50 & AP@75 \\
            \hline
            Vanilla Assign &  36.6 & 61.5 & 37.7 \\ \hline 
            Instance Assign & 34.2  & 60.2 & 34.9\\ \hline
            \end{tabular}
            \label{tab:ablation_inst}
\end{table}

\textbf{Instance-aware Feature Enhancement.} To achieve instance-aware feature enhancement, we introduce relative coordinates encoding. Specifically, we leverage a prior that proposals closer to a point annotation should have a higher chance to be assigned to this instance. Specifically, a feature map $f\in \mathcal{R}^{H \times W \times 2}$ is constructed for each instance to encode the relative coordinate offset to its point annotation $f_{ij}=[\Delta x_{ij}, \Delta y_{ij}]^T$, where $\Delta x_{ij}$ and $\Delta y_{ij}$ denote the coordinate offset of each pixel at index ${i, j}$ to the corresponding point annotation. The feature $f$ also follows the same RoI pooling procedure for each proposal and then gets concatenated with the normal RoI-pooled feature, as shown in Figure \ref{fig:pipeline}.  Due to the different coordinates of each group, instance-aware RoI features for different groups are generated even they have similar appearances, which is not uncommon in crowded scenes. 

\textbf{Instance-aware Parameter Generation.} We introduce instance-aware model parameters so that feature representations of proposals in different groups can be further diversified. Although a single proposal may have a larger IoU with the ground-truth bounding box of another group, for most proposals in a group, the maximum IoU is still achieved with the corresponding ground-truth instance. Thus, we calculate the mean of all RoI features in a group as the representation of the instance. More specifically, we first collect RoI features of all proposals in a group and apply spatial average pooling on them to produce vectorized features. Then, we compute the mean over these features to form the instance representation. Furthermore, to better leverage the category information of the point annotation, we introduce a class embedding matrix of shape $C \times 256$ where $C$ is the number of classes. This embedding matrix is optimized together with other model parameters. The embedding vector of the instance class and the constructed instance representation are concatenated, with which we produce instance-aware model parameters. Then we convolve the dynamically generated parameters with the above instance-aware proposal features to produce the classification score in the context of the instance-level assignment. This process is illustrated in the bottom-right component in Figure \ref{fig:pipeline} and summarized in Equation (\ref{eq:dgc}) where $0 \leq i \leq N-1$ and $0 \leq j \leq G-1$. $N$ and $G$ are the number of instances and size of groups respectively. $f^{i}_{j}$ is the instance-aware feature of j-th proposal in i-th instance group and $f'^{i}_{j}$ is the corresponding output feature of dynamic group convolution. $P_i$ is the generated parameter of the $i$-th group and $\mathbf{C}_i$ is the category embedding of the group $i$. $F(\cdot)$ stands for a linear projection with learnable parameters. The dynamic group convolution is a \textbf{high-order transformation} as the parameters are generated from averaged proposal features, which makes features from different groups more discriminative. 
\vspace{-2mm}
\begin{align}
    f'^{i}_{j} = f^{i}_{j} \otimes P_i \nonumber \\
    P_i = F(\frac{\sum_j f^{i}_{j}}{G}, \mathbf{C}_{i}) 
\label{eq:dgc}
\end{align}

The dynamic convolutions can be efficiently implemented as a group convolution where the number of groups is equal to the number of instances in the image. Through convolving instance-aware features with instance-aware parameters, we effectively increase the discrimination among instances. The generated convolution layer is used to process the RoI features of proposals in a certain group. The classification and regression heads are attached after our Dynamic Group Convolution to predict final class predictions and regression offsets.

\section{Experiments}
\label{sec:exp}
\textbf{Dataset.} We evaluate the performance of Group R-CNN on MS-COCO~\cite{lin2014microsoft} dataset. MS-COCO contains 118k training images with bounding box annotations and 5k images for validation. We randomly sample different percentages of images from the training set with fixed random seed 0 as our well-labeled set and use the rest as weakly-labeled sets with point annotation. To train our point-to-box regressor with well-labeled images, we randomly sample points within bounding boxes of instances at each iteration and use bounding boxes as the optimization target. After training, we run the inference process of the regressor on weakly-labeled images with point annotations. Following prior work~\cite{chen2021points}, we synthesize the point annotations for weakly-labeled images by randomly sampling a point from instance masks of objects for one-shot. We also evaluate our method on Pascal VOC~\cite{everingham2010pascal} and the results are in Appendix.

\textbf{Architecture Details.} Group R-CNN is built on top of Cascade R-CNN~\cite{cai2018cascade} and serves as the point-to-box regressor in WSSOD-P tasks.  To better leverage the category label of point annotations, we adopt class-aware detector RetinaNet~\cite{lin2017focal} as our region proposal network (RPN). Following the original RetinaNet, we use feature pyramid levels from $P_3$ to $P_7$ to extract region proposals. Notice that RetinaNet makes class-level predictions when extracting region proposals whereas the original RPN only produces objectness scores as in Faster R-CNN~\cite{ren2015faster}. For RoI pooling in the second stage, we drop $P_7$ and only use $P_3$ to $P_6$ as candidates.  For proposal refinement, a normal Cascade R-CNN head with three stages is used. Each stage has two shared FC layers and two separated FC layers to do the regression and classification, respectively. Unless otherwise specified, the model hyper-parameters are set as default~\cite{cai2018cascade}. RPN runs non-maximum suppression (NMS) with an IoU threshold of 0.7 on proposals from all groups with the same category based on the class prediction score and only 50 proposals are kept in each group. As for the final prediction results, we only select the proposal with the highest precision score in each group after NMS. In our dynamic group convolution, we always set the generated kernel size as $1$ for simplicity.

\textbf{Implementation Details.}
For a fair comparison, we follow the training setting of Point DETR~\cite{chen2021points} as much as possible. To train our Group R-CNN, we use 8 RTX 2080Ti GPUs with 2 images per GPU. Our data augmentation policy is also exactly the same as Point DETR. Though our proposed Group R-CNN has a different overall architecture with Point DETR~\cite{chen2021points}, we use ResNet-50~\cite{he2016deep} as our backbone, which is consistent with Point DETR. We use the learning rate of 0.02. We use SGD as our optimizer with a momentum of 0.9 and weight decay of $1e^{-4}$. The only difference in the training setting is the training schedule. Point DETR trains the regressor with 108 epochs and applies learning rate decay at the 72-th and 96-th epoch. \textbf{Our training schedule is substantially shorter than Point DETR}, we only train Group R-CNN by 50 epochs and decay the learning rate at the 30-th and 40-th epoch.

\subsection{Comparison with Point DETR}
\begin{figure}[t]
    \centering
    \includegraphics[width=\linewidth]{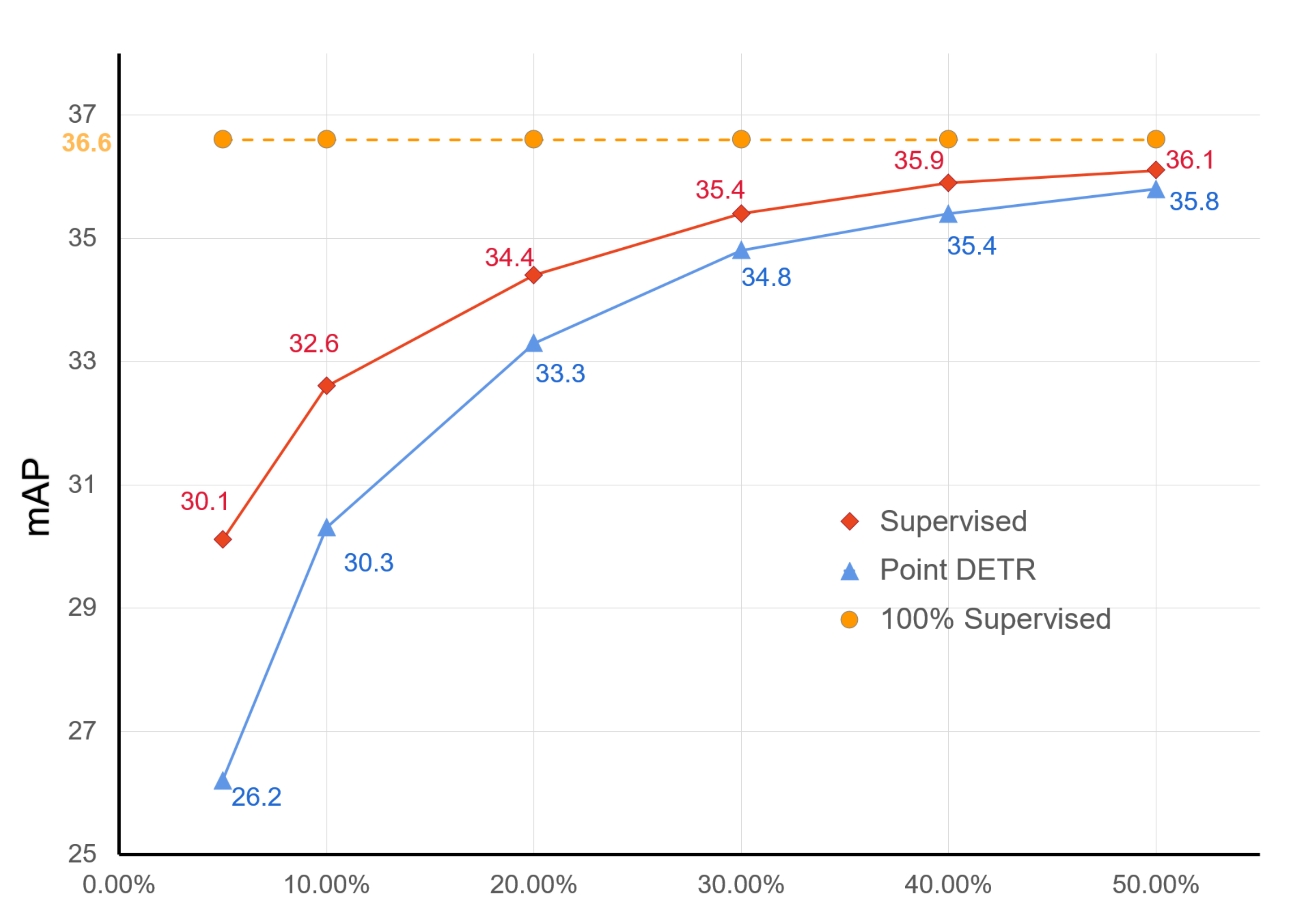}
    \caption{
    \textbf{Comparison with Point DETR.} Our method consistently outperforms Point DETR especially with limited well-labeled images. When using 50\% well-labeled images, the FCOS model trained with well-labeled images and pseudo-labeled images produced by our method closely matches the supervised baseline trained with 100\% well-labeled images.
    }
    \label{fig:main_results}
    \vspace{-5mm}

\end{figure}

We also follow the evaluation setting of Point DETR where we train our point-to-box regressor Group R-CNN with well-labeled images and inference on weakly-labeled images. As a final step, we train an FCOS detector with both well-labeled images and weakly-labeled images, which is also consistent with Point DETR. The training setting of FCOS follows the standard 1x training schedule with exactly the same hyper-parameters as in the standard supervised training setting. We report the performance of the FCOS detector in Figure \ref{fig:main_results}, which shows that Group R-CNN outperforms Point DETR with all different well-label fractions, especially when the well-labeled images are limited. We find that the fewer images are well labeled, the more Group R-CNN outperforms Point DETR. Specifically, in the most challenging scenario when only 5\% and 10\% images are labeled with bounding boxes, we achieve an improvement of 3.9 mAP and 2.3 mAP, respectively. Moreover, Group R-CNN only requires 50\% training epochs of Point DETR to achieve such a large improvement. When training with 50\% of well-labeled images, FCOS trained with pseudo-labeled images produced by Group R-CNN achieves comparable performance as the model trained with 100\% well-labeled images. It shows that we can largely close the gap between weakly semi-supervised detection and supervised detection with only point annotations.

\subsection{Ablation Study}

To illustrate the effectiveness of our proposed components, we conduct extensive ablation studies. In all of our ablation studies, Group R-CNN is trained with 10\% well-labeled images. To eliminate interference factors and demonstrate the effectiveness of our proposed components on the point-to-box regressor, we report test results on the COCO validation set with fixed point annotation. In other words, we only evaluate the performance of the point-to-box regressor instead of training an object detector with produced pseudo-labeled images. We train our Group R-CNN only with 24 epochs in this section because it is already sufficient to show the effectiveness of our design.

\begin{table}[htp!]
\centering
\caption{Impacts of sizes of different groups. Results are computed at IoU=0.5.}
\begin{tabular}{c|c|c|c|c}
\hline
 & AR & $AR_{s}$ & $AR_{m}$ & $AR_{l}$ \\
\hline
No Grouping & 80.4  & 61.1 & 86.8  & 95 \\
$k=1$ & 82.7  & 75.0 & 86.0 & 91.6 \\
$k=3$ & \textbf{90.6}  & 85.1 & 93.7 & 96.9\\
$k=6$ & \textbf{92.5} & 86.1 & 95.8 & 98.9\\
$k=9$ & \textbf{92.3} & 84.8  & 95.5  & 99.2\\
\hline
\end{tabular}
\label{tab:ablation_group}
\end{table}
\textbf{Size of Instance Group.} Recall that we select $k$ points around the projected annotation point on each level of the feature map and collect the output proposals as a group.  We conduct experiments with different $k$ values to study how the size of instance groups impacts the recall of our point-to-box regression. For the baseline with no proposal grouping, we run NMS on all proposals with IoU=0.7 and select the top-1000 proposals with the highest classification score. For proposal grouping, we keep 50 proposals for each group after NMS. On average, each image from COCO contains roughly 7.27 objects. Thus, our method produces $7.27 \times 50 = 363.5$ proposals as expectation. As shown in Table \ref{tab:ablation_group}, the performance of our RPN is significantly improved when instance grouping is adopted. With proposal grouping, although the model only produces around 30\% proposals compared with that of the model without instance grouping, the AR is improved by more than 10\%. Moreover, the performance is further boosted when $k$ increases from 1 to 3. Although a larger $k$ yields better results, $k > 3$ does not improve the recall by a large margin. Therefore, for a better performance-complexity trade-off, we choose $k=3$ as the default choice in Group R-CNN.

\begin{table}[t]
\caption{Ablation Study: Relative Coordinates}
\vspace{-3mm}
\centering
\begin{tabular}{c|c|c|c}
\hline
 & mAP & AP@50 & AP@75 \\
\hline
w/o relative coordinates & 35.7  & 61.0 & 37.0 \\
w/ relative coordinates & 37.1 & 64.1 & 38.1 \\
\hline
\end{tabular}
\label{tab:ablation_rc}
\end{table}

\textbf{Instance-Level Assignment.} Now we compare our instance-level proposal assignment with the vanilla proposal assignment. For the vanilla assignment strategy, we assign the proposals which have a maximal IoU with any ground-truth bounding boxes larger than predefined thresholds to be positive. The IoU thresholds are set as 0.5, 0.6, and 0.7 for the three stages, respectively, following the default setting as Cascade R-CNN~\cite{cai2018cascade}. NMS with an IoU threshold of 0.5 is applied at the last stage and only the proposal with the highest class score for each group is kept. For the instance-level assignment, we use the same IoU thresholds as the vanilla counterpart.

However, we find that directly applying instance-level assignment causes convergence difficulty. Although the R-CNN refinement stages are trained with the instance-level assignment. RPN is still trained by the vanilla assignment. Then RPN and R-CNN will have conflict requirements for the FPN. Therefore, we first detach FPN from the R-CNN heads to block the back-propagated gradients. Then we add additional projection layers on FPN to produce a separate FPN designated for R-CNN.

\begin{table}[t]
    \centering
    \caption{Impact of additional projection layers.}
    \vspace{-3mm}
           \begin{tabular}{c|c|c|c}
                \hline
                $\#$proj. layers & mAP & AP@50 & AP@75 \\
                \hline
                0 &  34.2 & 60.2 & 34.9 \\ \hline 
                1 &  \textbf{35.7} & 61.0 & 37.0 \\ \hline 
                2 & 35.8 & 60.9 & 36.9 \\ \hline 
                3 & 35.7  & 60.6 & 36.7 \\ \hline
                \end{tabular}
                \label{tab:ablation_projection}
\end{table}

We compare the effect of using different number of projection layers and the results are shown in Table \ref{tab:ablation_projection}. We show the effectiveness of additional projection convolution layers to process features with the detaching strategy. Using one additional projection layer lifts the mAP by 1.5 whereas more projection layers do not result in further improvements. Therefore, we simply use one additional projection layer so that the additional computation overheads are minimal.

\textbf{Effectiveness of Instance-aware Feature Enhancement.} As shown in Table \ref{tab:ablation_rc}, the mAP is improved from 35.7 to 37.1 when using relative coordinates. As discussed, although proposals from different instance groups may share similar appearances, enhancing the proposal feature with the relative coordinates endow the proposal with the ability to distinguish instances, given the prior that (ground truth) objects are located at different positions in the image.

\begin{table}[t]
    \centering
    \caption{Impact of Category Embedding (CE) and RoI Features (RoI-Feat.) on dynamic group convolution. In the baseline with 37.1 AP, all RoIs are not
refined by dynamic convolution.}
    \label{table:ablation_dgc}
    \begin{tabular}{cc|ccc}
    \hline
        \textbf{CE} & \textbf{RoI-Feat.} & mAP & AP@50 & AP@75 \\ \hline
                    &             &   \textbf{ 37.1}  &    64.1   & 38.1    \\
        \checkmark  &             &     38.5  &    64.7   & 39.9     \\ 
                    & \checkmark  &     38.2  &    64.5   & 39.7    \\
        \checkmark  & \checkmark  &     39.2  &    65.7   & 41.0    \\
    \hline
    \end{tabular}
\end{table}
\textbf{Effectiveness of Instance-aware Parameter Generation.} Table \ref{table:ablation_dgc} shows the results when generating convolution parameters with RoI features, category embeddings, and the concatenation of both, respectively. Clearly, generating parameters with both RoI features and category embeddings substantially outperforms using only a single component. Moreover, using a regular convolution layer in replacement of our dynamic group convolution only achieves 38.0 mAP (not included in the table), which is worse than dynamic group generation no matter what features are used in parameter generation. Therefore, our dynamic group convolution is a crucial design for Group R-CNN. We always set kernel size to 1 in all experiments since using larger kernel sizes such as 3 just gives marginal improvement (39.5 mAP for kernel size equals 3).

\begin{table}[t]
    \centering
    \caption{Comparison between \textbf{vanilla R-CNN with instance-level grouping} and \textbf{Group R-CNN} in crowded and non-crowded scenes, "vanilla" stands for the Vanilla R-CNN and "Group" denotes our proposed Group R-CNN.}

\begin{tabular}{c|c|ccc}
\hline
dataset                      & Method & mAP & AP@50 & AP@75 \\ \hline
\multirow{3}{*}{\begin{tabular}[c]{@{}c@{}}crowded\\ 54k images\end{tabular}}     & Vanilla    &  32.2   &   56.9    &   32.4    \\ 
                             & Group   &  35.4   &    62.5   &   35.8    \\ 
                             & $\Delta$   &  +3.2  &   +5.6  &  +3.4    \\ \hline
\multirow{3}{*}{\begin{tabular}[c]{@{}c@{}}non-crowded\\ 52k images \end{tabular}} & Vanilla    & 46.9    & 73.9      &  50.7    \\ 
                             & Group   &  48.5   & 75.0      &  52.7     \\ 
                             & $\Delta$   &  +1.6  &  +1.1   &  +2.0     \\ \hline
\end{tabular}
                \label{tab:ablation_crowded}
\vspace{-3mm}
\end{table}
\textbf{Compare with Vanilla Assignment.} In real-world datasets, it is very common that objects of the same class have overlaps. For example, more than 50\% images from MS-COCO contain crowded scenes. We compare the performance of Group R-CNN and vanilla R-CNN with instance-grouping under crowded scenes and non-crowded scenes. The models are evaluated on 90\% weakly-labeled images. Table \ref{tab:ablation_crowded} shows that Group R-CNN(with our proposed instance-level assignment, instance-aware feature enhancement and instance-aware parameter generation) significantly outperforms the baseline by 3.2 mAP and 1.6 mAP respectively. It is obvious that Group R-CNN performs particularly better in crowded scenes, which well-supports the motivation of our method.

\subsection{Visualization Results}
\begin{figure}[t]
    \centering
    \includegraphics[width=\linewidth]{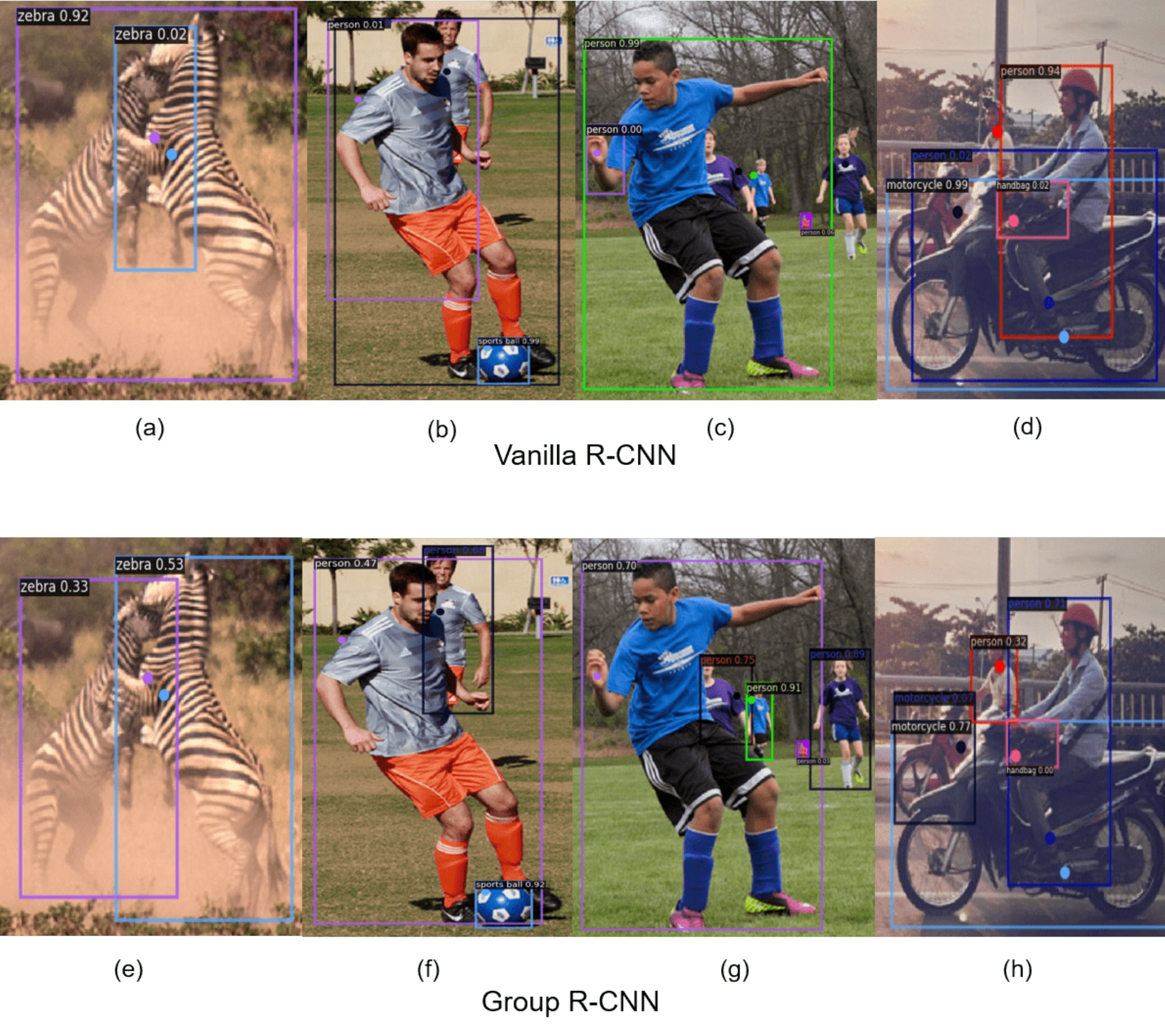}
    \caption{
    \textbf{Visualization of predicted pseudo bounding boxes by Group R-CNN (ours) and vanilla R-CNN.} Group R-CNN achieves high precision and recall in the point-to-box regression task ensuring one accurate bounding box is produced for each point annotation. In contrast, vanilla R-CNN fails to capture lots of instances even with point annotation.
    }
    \label{fig:vis}

\end{figure}

We provide qualitative analysis on the validation dataset and visualize the detection results of our point-to-box regressor Group R-CNN compared with vanilla R-CNN. As shown in Figure \ref{fig:vis}, each bounding box is generated with one point annotation (indicated by the same color). Vanilla R-CNN produces a number of bounding boxes that are largely overlapped with other instances resulting in a low recall rate. In contrast, Group R-CNN successfully produces bounding boxes for most point annotations even in crowded scenes with objects of the same category.

\section{Acknowledgement}
This project is supported by the Shanghai Committee of Science and Technology, China (Grant No. 20DZ1100800).
\section{Conclusion}
\label{sec:conclusion}
We propose Group R-CNN, a CNN-based point-to-box regressor for weakly semi-supervised object detection task. Group R-CNN leverages instance-level proposal grouping and instance-level representation learning(through instance-aware feature enhancement and instance-aware parameter generation) to improve the recall and precision. With these novel designs, Group R-CNN significantly outperforms existing transformer-based regressor by a large margin especially when well-labeled images are limited.

\clearpage
{\small
\bibliographystyle{ieee_fullname}
\bibliography{reference}
}
\clearpage

\appendix
In this supplementary material, we ask the following questions. Then we give answers to the above questions, one section for each question.

\begin{itemize}
  \item Why we choose to develop a CNN-based model rather than a transformer-based one?
  \item How does vanilla assignment (instead of instance-level assignment) work with the proposed instance-aware representation learning?
  \item How can Group R-CNN be improved with weakly-labeled images (only with point annotations)?
  \item To what extent does Group R-CNN outperform semi-supervised object detection methods?
  \item Can Group R-CNN generalize well on other benchmarks like VOC?
  \item What are the limitation and negative social impacts of Group R-CNN?
\end{itemize}

\section{Motivation: Inferior Performance of DETR When Data Lacks}
\begin{figure}[t]
    \centering
    \includegraphics[width=\linewidth]{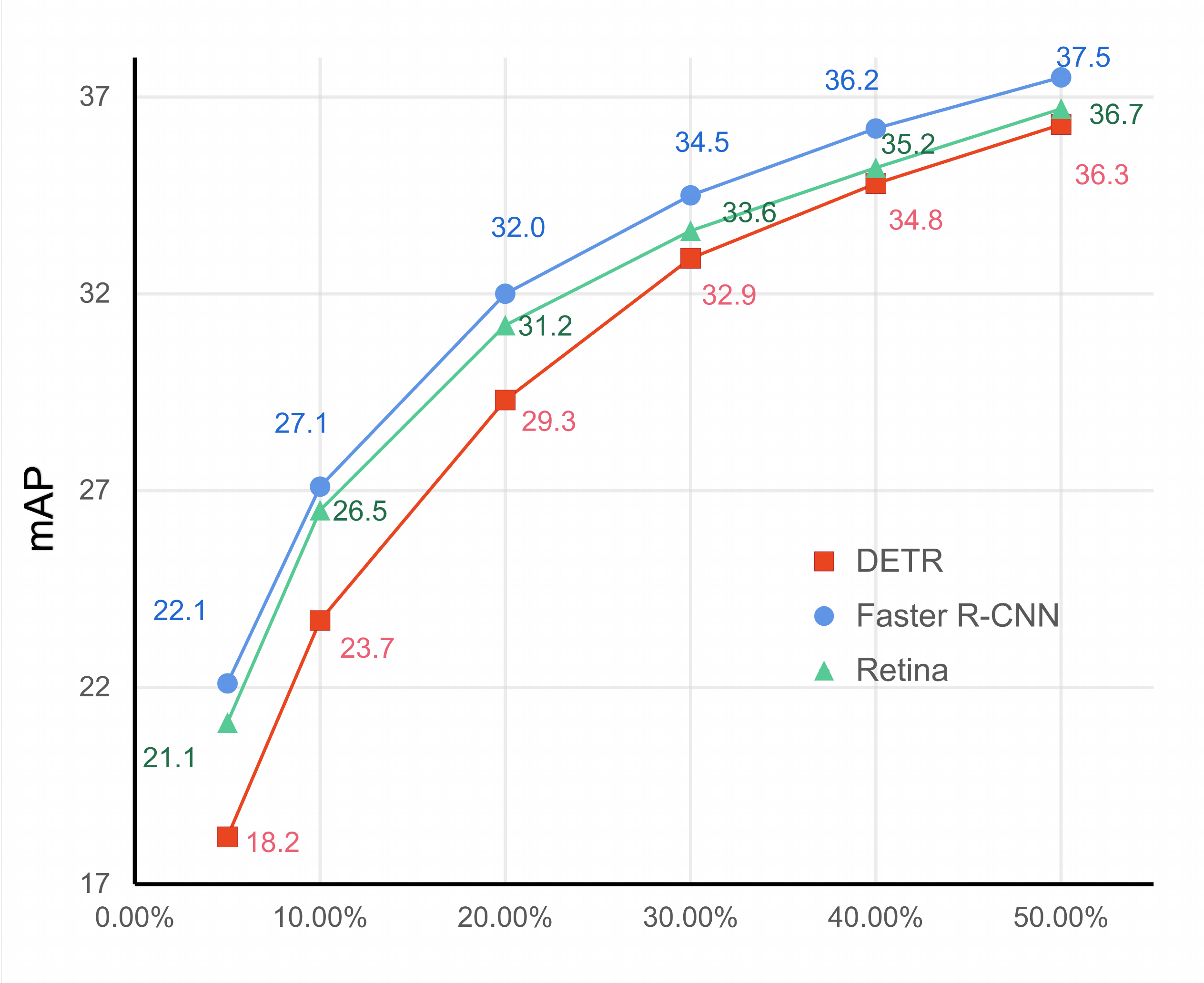}
    \caption{
    \textbf{Convergence Analysis of Different Detectors}
    }
    \label{fig:motivation}

\end{figure}

One of our motivations to develop a CNN-based model is that transformer-based models cannot generalize well when trained with insufficient data. We provide experimental results to support this argument. Specifically, we train two representative CNN-based object detectors (Faster R-CNN~\cite{ren2015faster} and RetinaNet~\cite{lin2017focal}) alongside a transformer-based detector (DETR~\cite{carion2020end}) with various percentages of images (labeled with bounding boxes) in a supervised fashion. They have similar performance when the entire COCO dataset is used. However, CNN-based detectors, especially the two-stage detector, are significantly better than DETR when training data is limited (see Figure \ref{fig:motivation}). The gap between CNN-based detectors and the transformer-based detector becomes larger when the number of labeled images decreases. It implies that the transformer-based model performs poorly in the case of data 
scarcity, but our pipeline involves training a point-to-box regressor with \textbf{limited data}, so it motivates us to develop a CNN-based regressor.

\begin{figure*}[htp!]
    \centering
    \includegraphics[scale=0.4]{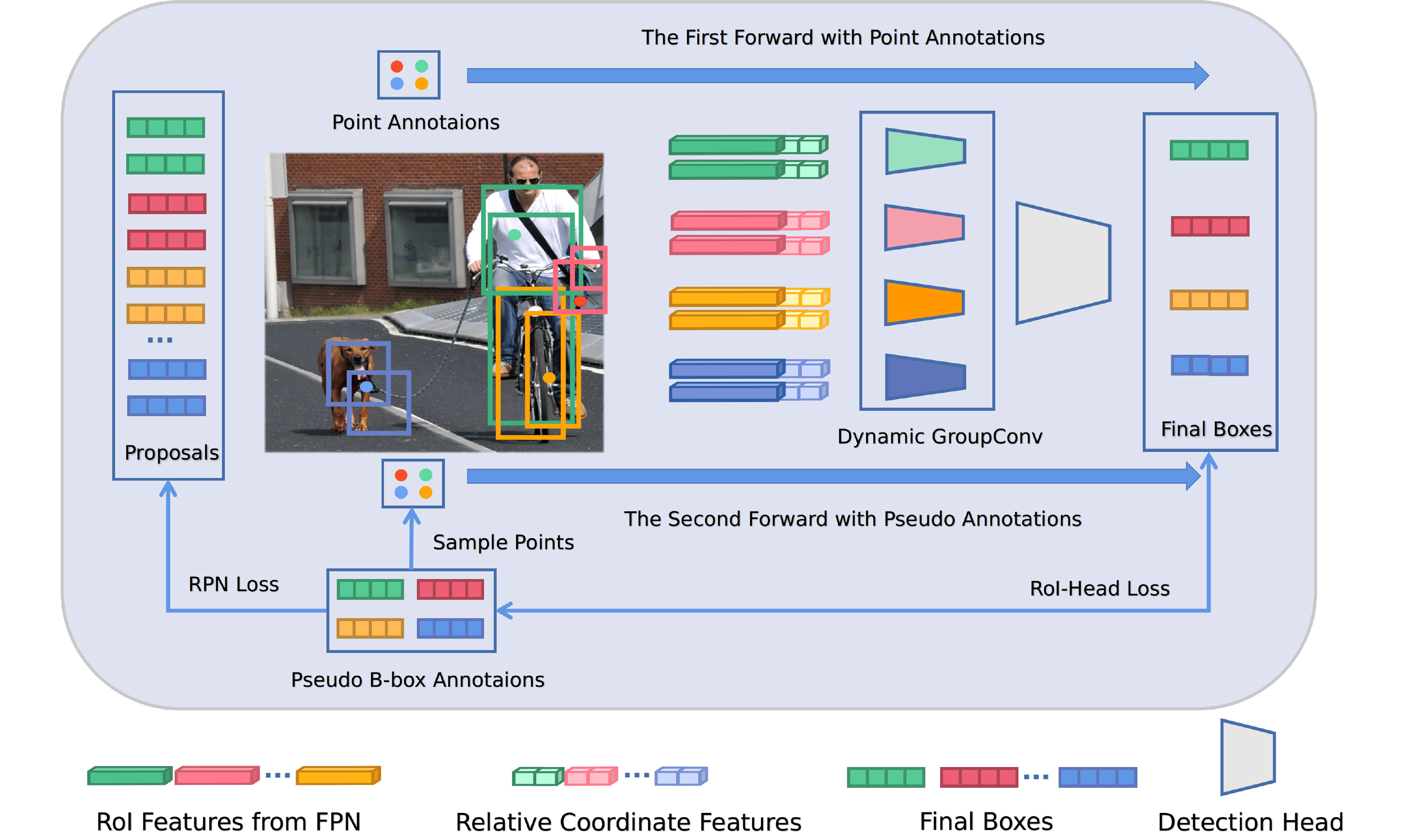}
    \caption{
    \textbf{The Pseudo-Labelling pipeline of Group R-CNN}. At each iteration, a weakly-labeled image is first forward to the model to obtain pseudo bounding box annotations. Next, a point is randomly sampled within each pseudo bounding box as pseudo point annotation. Finally, the model is trained with pseudo point annotations and pseudo annotations in the same fashion as training with well-labeled images.
    }
    \label{fig:semi_pipeline}

\end{figure*}
\section{Vanilla Assignment with Instance-aware Representation Learning}
In this work, we propose instance-aware feature enhancement and instance-aware parameter generation to comply with instance-level proposal assignment. We show that vanilla assignment is ineffective even with the above two strategies, demonstrating the necessity of instance-level assignment. We report the results of combining vanilla assignment with instance-aware feature enhancement and instance-aware parameter generation. Specifically, we evaluate the performance of the \textbf{baseline} (using instance grouping and vanilla assignment) when additionally incorporating: (1) 
detaching FPN~\cite{lin2017feature}, (2) additional projection convolution (PC, also in Table \textcolor{red}{4}, Section \textcolor{red}{4.2} of the main submission), (3) relative coordinates (RC) and (4) dynamic group convolution (DGC). For all experiments in this section, we exclude the default RoI sampling procedure to keep the size of instance group fixed to facilitate parallelization, so there is a slight performance drop (from 36.6 to 36.2) for Cascade R-CNN~\cite{cai2018cascade} when exploiting instance grouping.
\begin{table*}[t]
    \centering
    \caption{Using Vanilla Assignment Strategy in Group R-CNN}
           \begin{tabular}{c|c|c|c}
                \hline
                 & mAP & AP@50 & AP@75 \\
                \hline
                Casecade R-CNN + Instance Grouping &  36.2 & 60.9 & 37.6 \\ \hline 
                w/ deteaching &  34.5 & 59.4 & 35.4 \\ \hline 
                w/ 1 projection conv & 36.6 & 60.6 & 38.4 \\ \hline 
                w/ relative coordinates & 36.2  & 60.7 & 37.7 \\ \hline
                w/ 1 dynamic group convolution & 36.6 & 61.5 & 38.4 \\ \hline 
                \hline 
                Group R-CNN w/ vanilla assign & 37.2 & 61.6 & 38.8 \\ \hline
                Group R-CNN & \textbf{39.2}  &    \textbf{65.7}   & \textbf{41.0} \\ \hline 
                \end{tabular}
                \label{tab:cat_assign}
\end{table*}

Table \ref{tab:cat_assign} shows the results of adding each component to the baseline. It can be seen that most of our designs only bring marginal or no improvement over the baseline, and the detaching strategy even hurts the performance. When replacing instance-level assignment in Group R-CNN with the vanilla assignment and removing detaching (the last row in Table \ref{tab:cat_assign}), the model only achieves 37.2 mAP, which is significantly lower than Group R-CNN with the instance-level assignment (39.2 mAP). Hence, the instance-level assignment is an essential building block of Group R-CNN. 

\section{Improving Group R-CNN with Weakly-Labeled Images}

To compare fairly with Point DETR~\cite{chen2021points}, we follow their setting and train our point-to-box regressor Group R-CNN with only well-labeled images. However, our proposed framework is general and thus it is also possible to exploit weakly-labeled images during training, similar to semi-supervised learning (pseudo-labeling). The pipeline is shown in Figure \ref{fig:semi_pipeline}. At each iteration, for the weakly-labeled images, we first generate pseudo-bounding boxes with the Group R-CNN. Then, the generated boxes play the same role as those in well-labeled images. That said, we randomly sample a point within the pseudo bounding box to be the pseudo point annotation. Now the model can be trained with both weakly-labeled (with pseudo points as inputs and pseudo boxes as targets) and well-labeled images (with sampled points as inputs and human-labeled boxes as targets). We set the ratio between well-labeled images and weakly-labeled ones to 1:1 and the losses from weakly-labeled images are weighted by 0.5. We train Group R-CNN with the conventional multi-scale training for 50 epochs. Notice that we do not include any advanced strategies in the latest semi-supervised learning literature such as exponential moving average~\cite{tarvainen2017mean} and strong augmentation~\cite{sohn2020fixmatch}.

As shown in Table \ref{tab:semi}, training with both well-labeled and weakly-labeled images achieves a 2.6 mAP improvement. Even though our pseudo-labeling pipeline does not include any advanced designs in semi-supervised methods,
the results already show that such a pipeline is plausible and could yield better performance than our point-to-box regressor trained without using the weakly-labeled images (with \textbf{only} point annotations).
\begin{table}[htp!]
\caption{Training Group R-CNN in a Semi-Supervised Fashion}
\centering
\begin{tabular}{c|c|c|c}
\hline
 & mAP & AP@50 & AP@75 \\
\hline
w/o pseudo-labelling & 39.5  & 66.5 & 41.1 \\
w/ pseudo-labelling & 42.4 & 69.0 & 44.7 \\
\hline
\end{tabular}
\label{tab:semi}
\end{table}

\section{Comparing with Vanilla Semi-Supervised Learning}
In this work, we study weakly semi-supervised learning with point annotations. The key difference from vanilla semi-supervised learning is that instead of using \textbf{unlabeled} images without any form of annotations, we use \textbf{weakly-labeled} images with point annotations. We compare the performance of using weakly-labeled images and unlabeled images under STAC~\cite{sohn2020simple}, a semi-supervised object detection pipeline. STAC first trains a teacher model with only well-labeled images and produces fixed pseudo bounding boxes for unlabeled images. Then the pseudo bounding boxes are used to train a student model. We replace the pseudo bounding box produced by STAC with the ones generated by Group R-CNN to demonstrate the advantage of point annotations. To keep the fairness of comparison, we adopt the same strategy and hyperparameters (including augmentation, training steps, batch size et al.) as STAC when training the student model using the generated offline pseudo bounding boxes.  We evaluate the performance with 10\% well-labeled images from MS-COCO.

Using pseudo bounding boxes produced by Group R-CNN improves the student mAP by 5.4 mAP (from 28.7 to 34.1), which shows that the quality of pseudo bounding boxes produced by point annotations is significantly better than those generated directly from unlabeled images. The experimental results validate the efficacy of point annotations under the semi-supervised setting.

\section{Experiments on VOC datasets}
We achieve 76.2 AP
on VOC07+12 with 50 \% point annotation and 50\% box annotation. For reference, the AP of only 50\% boxes and
100\% boxes is 73.5\% and 77.4\%, respectively

\section{Limitation and Social Impact}
In this work, we follow the pipeline of Point DETR to tackle the problem of weakly semi-supervised object detection with points. Specifically, the pipeline involves: training a point-to-box regressor on well-labeled images, generating pseudo bounding boxes on weakly-labeled images, and training an object detector with the combination of well-labeled images and weakly-labeled images. The focus of this work is to develop a more accurate point-to-box regressor. However, when training such a regressor, only the well-labeled images are used. It is also possible to incorporate weakly-labeled images in the regressor training with a self-training fashion. 

The potential social impact of this work inherits from object detection. Annotation costs are significantly reduced with weakly point annotations but with our method, it is still possible to train a promising object detector with significantly lower labeling costs. Consequently, undesired applications of object detection systems such as surveillance may be more accessible. Note that any advances in object detection and low-label learning paradigm could result in similar social impacts. 
\clearpage
\end{document}